\documentclass[sigconf]{acmart}
\usepackage{subfig}
\AtBeginDocument{%
  \providecommand\BibTeX{{%
    \normalfont B\kern-0.5em{\scshape i\kern-0.25em b}\kern-0.8em\TeX}}}

\copyrightyear{2021} 
\acmYear{2021} 
\setcopyright{acmcopyright}\acmConference[PETRA 2021]{The 14th PErvasive Technologies Related to Assistive Environments Conference}{June 29-July 2, 2021}{Corfu, Greece}
\acmBooktitle{The 14th PErvasive Technologies Related to Assistive Environments Conference (PETRA 2021), June 29-July 2, 2021, Corfu, Greece}
\acmPrice{15.00}
\acmDOI{10.1145/3453892.3461009}
\acmISBN{978-1-4503-8792-7/21/06}

\begin{document}

\title{Sequential Late Fusion Technique for Multi-modal Sentiment Analysis}

\author{Debapriya Banerjee}

\orcid{0000-0001-6666-5863}
\affiliation{%
  \institution{The University of Texas at Arlington}
  \city{Arlington}
  \state{Texas}
  \country{USA}
  \postcode{76019}
}
\email{debapriya.banerjee2@mavs.uta.edu}

\author{Fotios Lygerakis}
\orcid{0000-0001-8044-3511}
\affiliation{%
  \institution{The University of Texas at Arlington}
  \city{Arlington}
  \state{Texas}
  \country{USA}}
\email{fotios.lygerakis@mavs.uta.edu}

\author{Fillia Makedon}
\affiliation{%
  \institution{The University of Texas at Arlington}
  \city{Arlington}
  \state{Texas}
  \country{USA}}
\email{makedon@uta.edu}

\renewcommand{\shortauthors}{Banerjee et al.}

\begin{abstract}
  Multi-modal sentiment analysis plays an important role for providing better interactive experiences to users. Each modality in multi-modal data can provide different viewpoints or reveal unique aspects of a user's emotional state. In this work, we use text, audio and visual modalities from MOSI dataset and we propose a novel fusion technique using a multi-head attention LSTM network. Finally, we perform a classification task and evaluate its performance.
\end{abstract}

\ccsdesc[500]{Computing methodologies~Modeling methodologies}


\keywords{late fusion, multi-modal sentiment analysis, multi head attention recurrent neural networks}

\maketitle

\section{Introduction}
Social media is an industry that relies on efficient emotion recognition and sentiment analysis (SA), in order to provide relevant content to the users by exploiting their preferences and habits  \cite{BALAZS201695}.
Furthermore, the continuously increasing popularity of social robots is another eminent field that can be greatly benefited by tracking user's mental state, offering more organic and engaging human-robot interaction \cite{nao} or user-specific services \cite{social_robots}.

\section{Related Work}
The ever-growing user-provided content on social media consisting of different modalities, i.e., audio, text and video, has already created huge dataset \cite{ MOSI}, that eased the utilization of neural network (NN) techniques on these fields. Literature has shown that multi-head attention recurrent NN (MHA-RNN) in combination with late fusion (LF) \cite{MuSe-Car, 10.1145/3372278.3390698} approaches are the most prominent techniques. As discussed in  \cite{MAJUMDER2018124} by N.Majumder et al., sequential (or hierarchical in their case) late fusion can filter out inter-modal correlation.

In most recent approaches, an off-the-shelf encoder is used for each modality. Then the encoding of each modality is being fused by an  RNN. In \cite{10.1145/3372278.3390698}, the authors use word2vec \cite{mikolov2013efficient}, OpenSmile \cite{10.1145/1873951.1874246} and 3D Convolutional NN (CNN) \cite{dd6ac3cf18454e6d97f480f126701ab7} for text, audio and visual modalities respectively, the encoding of which they use in a bidirectional MHA-RNN and a softmax layer at the end to perform binary sentiment classification. A similar approach is followed by Poria et al. in  \cite{8636432}, but with the exception of using a custom CNN for the text encoding and a support vector machine (SVM) for the classification task. 

In this paper, we propose a LF technique that uses the same uni-modal encoders as in  \cite{10.1145/3372278.3390698}. Our technique is also based on an MHA-RNN model, however, the LF technique we use takes advantage of the sequential temporal dependencies across the modality encodings. 

\section{Method}


\subsection{Uni-modal Feature Extraction}

In this work, we use LF combining uni-modal features extracted from 2199 short video utterances of the MOSI dataset \cite{zadeh2016mosi}. The utterances were annotated with sentiment labels ranging from -3 (very negative sentiment) to +3 (very positive sentiment) and were produced by 89 speakers between 20-30 years old.
From the video dataset, we first extract the audio and then  convert it into text using IBM Watson, a Speech to Text converter \footnote{https://speech-to-text-demo.ng.bluemix.net}.
Each modality consists of sequential utterances that are fed into three different modality-specific encoders to extract the visual, audio and textual uni-modal features. For textual features we use the text-CNN model \cite{mikolov2013efficient}, for audio features we use OpenSmile  \cite{10.1145/1873951.1874246} and for the visual ones 3D-CNN model  \cite{dd6ac3cf18454e6d97f480f126701ab7}. All the above uni-modal extractors provide 300-dimensional features.

\subsection{Fusion Technique}

\begin{figure}[h]
  \centering
  \includegraphics[width=0.50\textwidth]{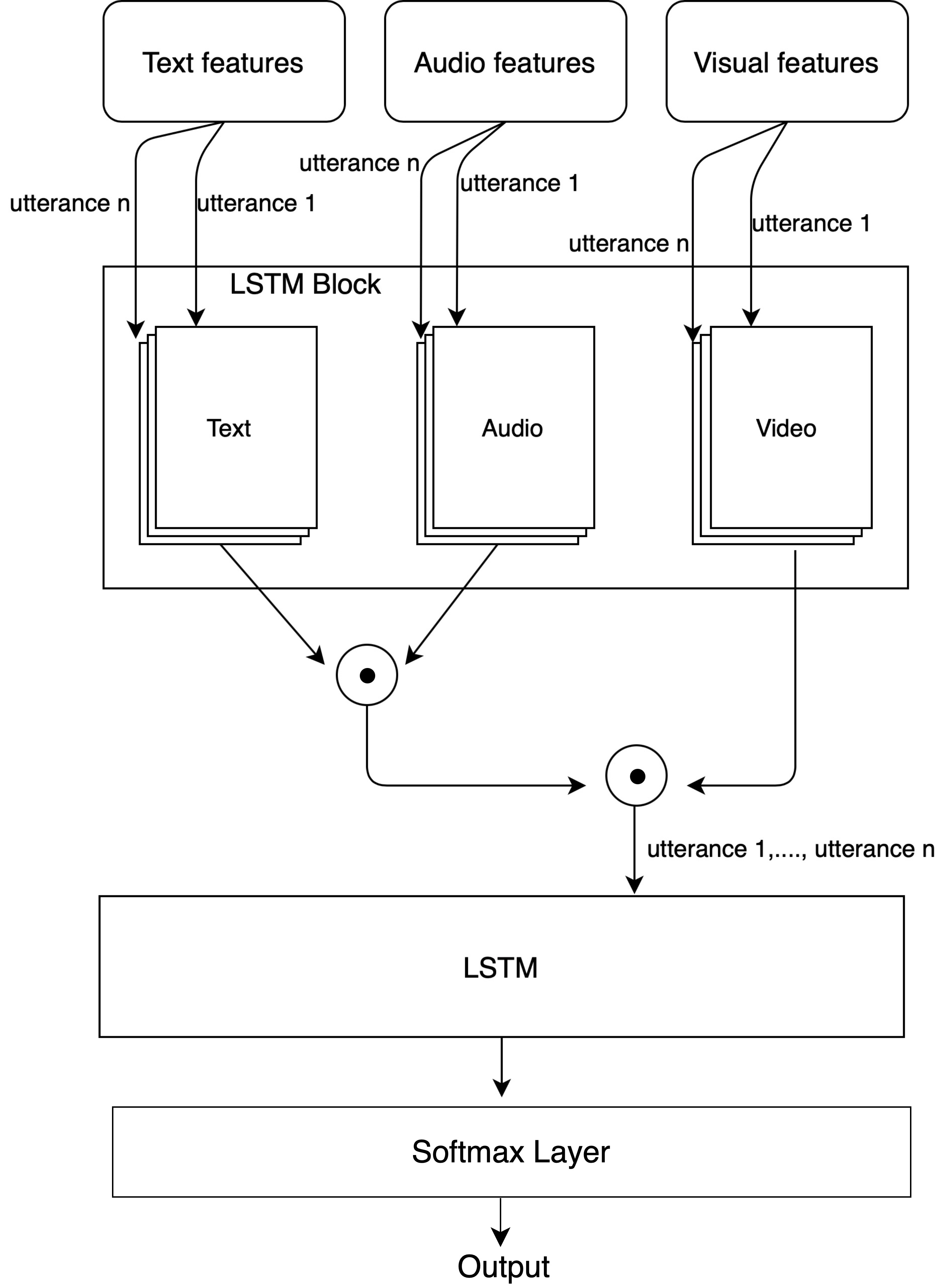}
  \caption{Fusion architecture using MHA-LSTM}
\label{fusion2}  
\end{figure}

Our LF approach consists of two layers of LSTM nodes with multi-head attention (MHA-LSTM). The first layer (LSTM Block in Figure \ref{fusion2}) contains three nodes, one for each of the three modalities, i.e. each node takes as input a sequence of utterances of a particular modality from textual, audio or visual features. We fuse the states of the first layer LSTM nodes in two steps: 

1. First, we fuse the states of the LSTM nodes that correspond to the textual and audio features by calculating their inner product.

2. Secondly, we calculate the inner product of the above outcome with the state of the LSTM node that corresponds to the visual features. 

We observed that fusing textual and audio features first gives better performance. Consequently, the product of the second step is given as input to the second layer of our MHA-LSTM network (LSTM in Figure \ref{fusion2}). Finally, the output of the MHA-LSTM network is given to a softmax layer which classifies among the 7 different review categories of the dataset.



   


\subsection{Experiments}
We train the MHA-LSTM network using the categorical cross-entropy as a loss function and we optimize using ADAM. Our model converges after 100 epochs. We split the dataset into 80\% of training samples and 20\% of testing ones. We evaluate our model in terms of accuracy, precision, recall and F1 score. Table \ref{tab:c1} shows the detailed evaluation metrics.

\begin{table}[H]
\caption {\textbf{Performance metrics on our proposed model with MOSI dataset}}
\begin{tabular}{|l|l|}
\hline
\multicolumn{1}{|c|}{\textbf{Metric}}         & \multicolumn{1}{c|}{\textbf{Score}} \\ \hline
Accuracy                               & 0.769                                               \\ \hline
Precision   & 0.837
                               \\ \hline
Recall              & 0.781                                 \\ \hline
F1 Score                          & 0.808               \\ \hline  
\end{tabular}
\label{tab:c1}
\end{table}


\section{Discussion and Future Work}

In this work, we propose a novel multi-modal SA method using LF with an MHA-LSTM model on uni-modal off-the-shelf feature extraction techniques. Our intention is to explore the potential of a different LF technique using a multi-head attention architecture.
In the future, we plan to perform more extensive experiments with other datasets on SA exploiting again multiple modalities. These datasets will also involve other areas of sentiment like Post-Traumatic Stress Disorder (PTSD) and general stress.

\bibliographystyle{ACM-Reference-Format}
\bibliography{sample-base}

\end{document}